\documentclass[a4paper,fleqn]{cas-dc}

\usepackage[authoryear,round]{natbib}

\setcitestyle{aysep={},yysep={;}}

\usepackage{amsfonts}

\def\tsc#1{\csdef{#1}{\textsc{\lowercase{#1}}\xspace}}
\tsc{WGM}
\tsc{QE}

\begin{document}
\let\WriteBookmarks\relax
\def\floatpagepagefraction{1}
\def\textpagefraction{.001}

\shorttitle{Survey: Transformer based Video-Language Pre-training}    

\shortauthors{L Ruan et al.}  

\title [mode = title]{Survey: Transformer based Video-Language Pre-training}  



%

\author[]{Ludan Ruan}{}



\ead{ruanld@ruc.edu.cn}



\affiliation[]{organization={School of Information, Renmin University of China},
            city={Beijing},
            country={China}}

\author[]{Qin Jin}

\fnmark[*]
\ead{qinj@ruc.edu.cn}




\cortext[1]{Corresponding author}



\begin{abstract}
Inspired by the success of transformer-based pre-training methods on natural language tasks and further computer vision tasks, researchers have begun to apply transformer to video processing. 
This survey aims to give a comprehensive overview on transformer-based pre-training methods for Video-Language learning. 
We first briefly introduce the transformer structure as the background knowledge, including attention mechanism, position encoding etc.
We then describe the typical paradigm of pre-training \& fine-tuning on Video-Language processing in terms of proxy tasks, downstream tasks and commonly used video datasets.
Next, we categorize transformer models into Single-Stream and Multi-Stream structures, highlight their innovations and compare their performances.
Finally, we analyze and discuss the current challenges and possible future research directions for Video-Language pre-training.

\end{abstract}



\begin{keywords}
Transformer\sep Muti-modal Pre-training\sep Video-Language Pre-training
\end{keywords}
\maketitle

\section{Introduction}
Transformer networks~\citep{Vaswani_Transformer_Nips17} have shown their great advantage on performance and become popular in Deep Learning~(DL).
Compared to traditional deep learning networks such as Multi-Layer Perceptrons~(MLP), Convolutional Neural Networks~(CNNs) and Recurrent Neural Networks~(RNNs), transformer is more suitable for pre-training \& finetuing, because its network structure is easy to deepen and its smaller model bias. 
The typical pre-training \& fine-tuing paradigm is that the model is first trained on a large amount of (typically self-supervised) training data and then fine-tuned on smaller (typically task specific) datasets for the downstream tasks. The pre-training stage helps the model to learn the universal representation, which benefits downstream tasks.
 
Transformer based pre-training method was first proposed for Natural Language Processing~(NLP) tasks and achieved remarkable performance gains. For example, Vaswani et al.~\citep{Vaswani_Transformer_Nips17} firstly propose the transformer structure with self-attention mechanism for machine translation and English constituency parsing tasks. BERT - Bidirectional
Encoder Representations~\citep{Devlin_BERT_coRR18} can be considered as a milestone in NLP, which adopts the transformer network for pre-training on unlabeled text corpus and achieves the state-of-the-art performance on 11 downstream tasks.  GPT - Generative Pre-trained Transformer v1-3~\citep{Radford_GPT1_2018,Radford_GPT2_2018,Brown_GPT3_NeurIPs20} are designed as general language models with extended parameters and trained on extended training data, among which GPT-3 is trained on  45TB of compressed plain text data with  175 billion parameters.
Inspired by the breakthrough of transformer based pre-training methods in the NLP field, researchers in computer vision~(CV)  have also applied transformers in varies tasks in recent years.
For example, DETR~\citep{Carion_DETR_ECCV20} removes the bounding box generation stage for object detection based on transformer networks. Dosovitskiy et al.~\citep{Dosovitskiy_ViT_ICLR21} apply a pure transformer ViT that directly handles sequences of image patches and proves its effectiveness for image classification based on large training set.

Video analysis and understanding is more challenging, because video naturally carries multi-modal information. 
For the representative Video-Language tasks such as video captioning~\citep{Das_VideoCap_CVPR13} and video retrieval~\citep{Xu_MSRVTT_CVPR16}, existing methods have mainly focused on learning video's semantic representation based on the video frame sequence and corresponding captions.
In this paper, we focus on providing a comprehensive overview of the recent advances in transformer based pre-training methods for Video-Language processing, including commonly used metrics of corresponding benchmarks, taxonomy of existing model designs, and some further discussion. We hope to track the progress of this area and provide an introductory summary of related works for peer researchers, especially beginners.

The remainder of this paper is organized as follows: 
Section 2 introduces the related fundamental concepts, including standard transformer with self-attention mechanism, the paradigm of pre-training \& finetuning approach, and commonly used datasets. 
Section 3 presents the major existing methods according to their model structures and highlights their strength and weakness as well. 
Section 4 further discusses several research directions and challenges, and Section 5 concludes the survey.

\begin{figure*}[t]
  \centering
  \includegraphics[width=\textwidth]{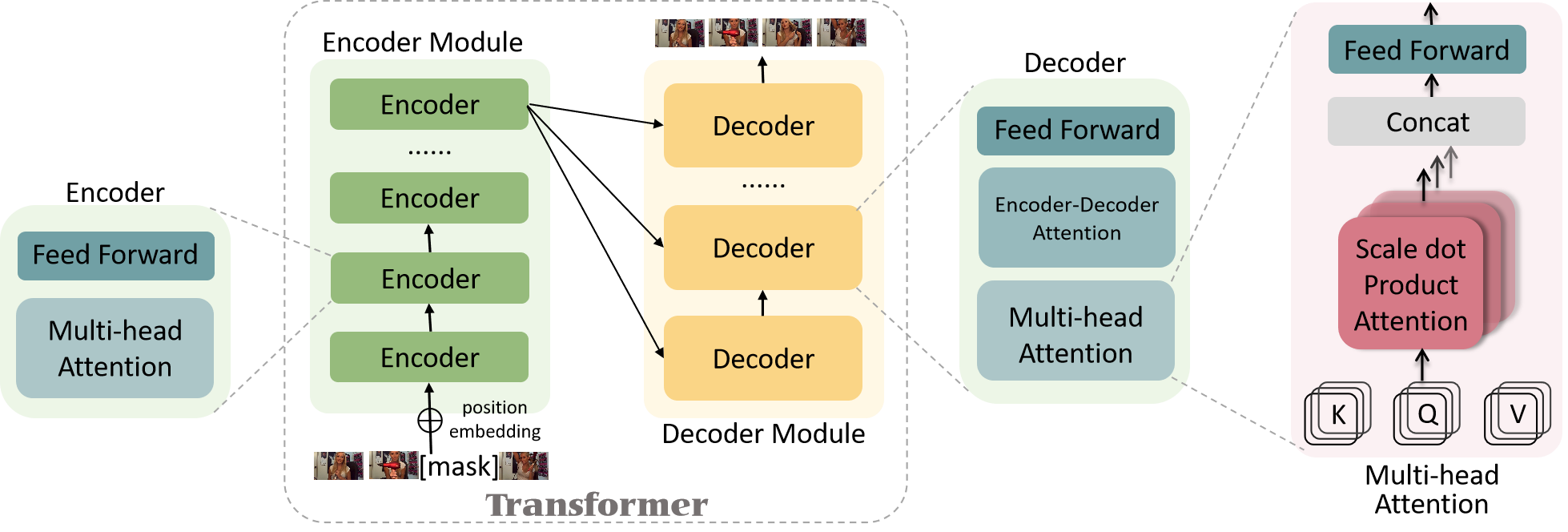}
  \caption{An overview of the standard transformer architecture. The whole transformer is composed of encoder module and decoder module, with several encoders and decoders stacked in each module respectively. Each encoder consists of a multi-head attention layer and a feed forward layer, while each decoder additionally contains a encoder-decoder attention layer. The multi-head attention mechanism is shown in the right most column, which transfers the input sequence into $h$ groups of \{K, Q, V\} and concatenates the self-attention outputs of each group as the final output.}
  \label{fig:transformer_structure}
\end{figure*}

\section{Background Fundamentals}

\subsection{Transformer}
Transformer~\citep{Vaswani_Transformer_Nips17} was first proposed in the field of Neural Language Processing (NLP) and showed great performance on various tasks~\citep{Wang_Glue_EMNLP18,Rajpurkar_SQUAD_EMNLP16,Zellers_SWAG_EMNLP18}. It has been successfully applied in other fields ever since, from language~\citep{Devlin_BERT_coRR18, WRae_GPT_ICLR20} to vision~\citep{Dosovitskiy_ViT_ICLR21}. 


As illustrated in Fig.~\ref{fig:transformer_structure}, the standard transformer consists of several encoder blocks and decoder blocks. Each encoder block contains a self-attention layer and a feed forward layer, while each decoder block contains an encoder-decoder attention layer in addition to the self-attention and feed forward layers.

\subsubsection{Self-Attention}
Self-attention is one of the core mechanisms of transformer,
which exists in both encoder and decoder blocks. 
Taking a sequence of entity tokens $X=\{x_0, x_1, ..., x_n\}$ as input~(the entity tokens can be word sequence in NLP or video clips in the vision area), self-attention layer first linearly transforms the input tokens into three different vectors: key vector $K \in \mathbb{R}^{n \times d^K}$, query vector $Q \in \mathbb{R}^{n \times d^Q}$ and value vector $V \in \mathbb{R}^{n \times d^V}$~(e.g. $d^K=d^Q=d^V=512 $ in practice).
The output is produced via $\text{Att}(X)=\text{softmax}(\frac{Q \cdot K^T}{\sqrt{d^Q}} \times V)$
where $Q \cdot K^T $ is to capture the relevance score between different entities,  $\sqrt{d^Q}$ is to reduce the score for gradient stability, $\text{softmax}$ operation is to normalize the result for probability distribution and finally, multiplying with V is to obtain the weighted value matrix.

In the decoder block, the encoder-decoder attention is similar to self-attention, with the key vector $K$ and the query vector $Q$ from encoder module and the value vector $V$ from the output of the previous decoder block. 

Note that not all self-attention attend to all entities. In the training stage of BERT~\citep{Devlin_BERT_coRR18}, 15\% of the input tokens are randomly masked for prediction and the masked entities should not be attended. When using BERT to output the next word token in the downstream task of sentence generation, the self-attention layer of decoder block only attends to the previous generated entities. 
Such attention can be realized by a mask $M \in \mathbb{R}^{n \times n}$, where the corresponding masked position of M is set zero. The formula of masked self-attention can be adjusted from the original self-attention to $\text{MaskedAtt}(X)=\text{softmax}(\frac{Q \cdot K^T}{\sqrt{d^Q}} \circ M)\times V$.

\subsubsection{Multi-Head Attention}
Multi-head attention mechanism~\citep{Vaswani_Transformer_Nips17} has been proposed to model the complex relationships of token entities from different aspects. To be specific, the input sequence $X$ is linear transformed into $h$ groups of $\{K_i,Q_i,V_i\}_{i=0}^{h-1}$, each group repeats the self-attention process. The final output is produced by projecting the concatenation of the outputs from the $h$ groups with a weight matrix $W \in \mathbb{R}^{h d^V \times d}$. The overall process can be described as:
\begin{equation*}
\begin{aligned}
    \text{MultiHeadAtt}(X)&=\left[ \text{Att}_0(X),\text{Att}_1(X),...,\text{Att}_{h-1}(X)\right]W\\
    \text{Att}_i(X)&=\text{softmax}(\frac{Q_i \cdot K_i^T}{\sqrt{d^Q_i}}) \times V_i
\end{aligned}
\end{equation*}

\subsubsection{Position Encoding}

Different from CNNs~\citep{Y_CNN_1998} or RNNs~\citep{Chung_GRU_2014}, self-attention lacks the ability to capture the order information of the sequence. To address this problem, position encoding~\citep{Vaswani_Transformer_Nips17} is added to the input embedding in both the encoder and decoder blocks. 
The position encoding of tokens are constructed as follows:
\begin{equation*}
\begin{split}
    \text{PE}(pos,2i)=sin(\frac{pos}{10000^{2i/d_{model}}})\\
    \text{PE}(pos,2i+1)=cos(\frac{pos}{10000^{2i/d_{model}}})
\end{split}
\end{equation*}
where $pos$ refers to the token's position and $i$ refers to the dimension. 
Another commonly used way to introduce position information is learned position embedding~\citep{Gehring_Seq2Seq_ICML17}. Experiments in~\citep{Vaswani_Transformer_Nips17} show that these two position encoding methods achieve similar performance.

\subsubsection{Transformer Structure}
The original Transformer~\citep{Vaswani_Transformer_Nips17} follows the encoder-decoder structure with stacks of 6 encoder blocks and 6 decoder blocks respectively. The encoder block consists of a multi-head self-attention sub-layer and a position-wise feed-forward sub-layer, where the position-wise feed-forward sub-layer contains two linear transformations with a ReLU activation. The decoder block additionally inserts a third sub-layer of encoder-decoder attention.  What's more, residual connection and layer normalization is added to each single block for further performance promotion. All sub-layers in the model, as well as the embedding layers, produce outputs of dimension $d_{model}$ = 512, and the  dimension of hidden layer is $d_h$ = 2048. 

Compared with CNNs and RNNs, the major advantages of transformer are the ability to simultaneously capture global information and parallel computation. Furthermore, the concise and stackable architecture of transformer enables training on larger datasets, which promotes the development of \textit{pre-training} \& \textit{fine-tuning} self-supervised learning.

\subsection{Pre-training \& Fine-tuning}
Pre-training \& Fine-tuning has become a typical learning paradigm for transformer based models:  
first pre-training model on large-scale dataset in supervised or unsupervised way and then adapting the pre-trained model on smaller datasets for specific downstream tasks via fine-tuning. Such paradigm can avoid training new models from scratch for different tasks or datasets. It has been proved that pre-training on larger datasets helps learning universal representations, which improves the performance of downstream tasks.
For example, NLP Transformer model GPT~\citep{Radford_GPT1_2018} gains average 10\% absolute improvement on 9 downstream benchmark datasets~(e.g. CoLA~\citep{warstadt_CoLA_2018}, MRPC~\citep{Dolan_MRPC_IJNLP05}) after pre-training on BooksCorpus dataset~\citep{Zhu_BooksCorpus_ICCV15} with 7000 unpublished books.
Vision Transformer model ViT-L/32~\citep{Dosovitskiy_ViT_ICLR21} gains 13\% absolute accuracy improvement on the test set of ImageNet~\citep{Deng_Imagenet_2009}  after pre-training on JFT-300M~\citep{Sun_JFT300_ICCV17} with 300 million images.

Owing to the successful application of pre-trained models in  NLP and CV tasks, more and more researches explore the  cross-modal tasks, including Vision-Language and Video-Language. The main difference between Vision-Language tasks and Video-Language tasks is that the former focus on the image and text modalities such as language based image retrieval~\citep{Lee_imgretr_ECCV18} and image captioning~\citep{Vinyals_Caption_CVPR15}, while the later  focuses on the video and text modalities, which adds the temporal dimension over the image modality. 

In following subsections, we describe the Pre-training \& Fine-tuning methods in Video-Language field, including the commonly used proxy tasks and video-language downstream tasks.

\subsubsection{Proxy Tasks}
\label{section:proxy_tasks}
Proxy tasks are crucial for the final performance of pre-trained models as they directly determine the models' learning objectives. 
We classify the proxy tasks into three
categories: Completion tasks, Matching tasks and Ordering tasks.
1) \textit{Completion tasks} aim to reconstruct the masked tokens of input, which endow the model with the ability of building intra-modal or inter-modal relationships. Typical tasks include Masked Language Modeling~(MLM), Masked Frame Modeling~(MFM), Masked Token Modeling~(MTM), Masked Modal Modeling~(MMM) and Language Reconstruction~(LR). We will describe them in details in the following section.
2) \textit{Matching tasks} are designed to learn the alignment between different modalities, originating from Next Sentence Prediction~(NSP) of BERT~\citep{Devlin_BERT_coRR18}. For example, Video Language Matching~(VLM) is the classical matching task, which aims at matching video and text modalities. Some researchers also introduce the audio modality for further matching objective~\citep{Akbari_VATT_arxiv21}.
3) \textit{Ordering tasks} are to shuffle the sequence at the input side and force the model to recognize the original sequence order. For example, Frame Ordering Modeling~(FOM) is specifically designed to exploit the temporal nature of video sequence and Sentence Ordering Modeling~(SOM) is designed for the text modality.

Among all commonly used proxy tasks, \textit{Self-Supervised Learning}~(SSL) is the dominant strategy adopted in order to adapt to the situation that pre-training  requires massive  training data. SSL is one type of \textit{UnSupervised Learning}~(USL) that generates labeled data automatically itself, which inspires the model to learn the inherent co-occurrence relationships of data. 
For example, in the sentence completion task such as ``I like \_\_\_ books'', a well-trained language model should fill in the blank with the word ``reading''.
In Video-Language pre-training, Masked Language Modeling~(MLM) and Mask Frame Modeling~(MFM) are two widely used SSL proxy tasks .



Contrastive Learning~(CL)~\citep{Chen_CL_ICML20} has recently become an important component in self-supervised learning for Video-Language pre-training. Different from generating masked tokens with  measuring L2 distance, it embeds the
same samples close to each other while trying to push away the embeddings from different samples. An extensive survey of CL can be found in~\citep{Jaiswal_CL_2020}.

In the remainder of this section, we introduce some widely used proxy tasks (as summarized in Tab.~\ref{tab:proxy_tasks}) during Video-Language pre-training. For the following formulas, we use the general notations of $w, v, t$ as word sequence, video sequence and the union tokens of $v$ and $w$. $w_m, v_m, t_m$ refer to the corresponding masked tokes.

\paragraph{\textbf{Masked Language Modeling~(MLM)}} 
was first referred to as a cloze task in~\citep{L_MLM_1953} and then adapted as a proxy task during the pre-training of BERT~\citep{Devlin_BERT_coRR18}. Original MLM is to randomly mask out a fixed percentage of words from the input sentence, and then predict the masked words based on other word tokens. MLM used in Video-Language pre-training not only learns the inherent co-occurrence relationships of sentence but also combines the visual information with the sentence. For example, as elaborated in ActBERT~\citep{Zhu_ActBERT_CVPR20}, when a verb is masked out, the task forces the model to extract relevant action features for more accurate prediction. When a noun or a description of noun is masked out, visual features of related object can provide contextual information.
Empirically, the masking percentage is always set 15\%. The loss function of MLM can be defined as:
\begin{equation*}
    \mathcal{L}_\textit{MLM} = -\mathbb{E}_{w_m \sim w}(logP(w_m|w_{\backslash w_m},v))
\end{equation*}

\paragraph{\textbf{Masked Frame Modeling~(MFM)}} 
is similar to MLM in that it simply changes the sentence to the video sequence.
That is, the frame tokens are masked for prediction according to the contextual frames and the input text for semantic constraints.

However, since a video is continuous, with no fixed vocabulary as text, researchers make different adjustments on the input side or loss objective side for the MFM task. We categorize MFM into three sub tasks according to loss functions: 1) MFM with Cross Entropy~(MFMCE), 2) MFM with Regression~(MFMR), and 3) MFM with Contrastive Learning~(MFMCL).

The typical examples of MFMCL can be found in VideoBERT \citep{Sun_videoBERT_ICCV19} and ActBERT \citep{Zhu_ActBERT_CVPR20}. VideoBERT splits the continuous videos into clip tokens and clusters clip tokens into a fixed size of dictionary by hierarchical k-means. In this way, the masked video feature can be predicted as video word with class likelihood. ActBERT extracts the action concept and local object feature from the video and the model is forced to predict the action  category and object category of masked video tokens respectively. The loss function of MFMCL can be defined as:
\begin{equation*}
    \mathcal{L}_\textit{MFMCL} = -\mathbb{E}_{v_m \sim v}(logP(v_m|v_{\backslash v_m},w))
\end{equation*}

The typical examples of MFMR can be found in HERO \citep{Li_HERO_EMNLP20}, which learns to regress the output on each masked frame $v_m$ to its visual features. HERO uses L2 regression between the input video feature $v_m$ and the output video feature $h(v_m)$:
\begin{equation*}
    \mathcal{L}_\textit{MFMR} = \mathbb{E}_{v_m \sim v}(||h(v_m)-v_m||^2)
\end{equation*}

However, it is hard to reconstruct the original video feature with regression as a video contains rich information. MFMCO adapts Contrastive Learning~\citep{Chen_CL_ICML20} to maximize the Mutual
 Information~(MI) between the masked video tokens and the original video tokens:
 
 \begin{equation*}
 \begin{aligned}
    & \mathcal{L}_\textit{MFMCO} = -\mathbb{E}_{v_m \sim v}(log\text{NCE}(v_m|v_{\backslash v_m},w))\\
    & \text{NCE}(v_m|v_{\backslash v_m},w) = \frac{\text{exp}(h(v_m)v_m^T)}{\mathcal{Z}} \\
    & \mathcal{Z} = \text{exp}(h(v_m)v_m^T)+\sum_{v_j \in v_{\backslash v_m}}\text{exp}(h(v_m)v_j^T)
 \end{aligned}
 \end{equation*}
 
\paragraph{\textbf{Masked Token Modeling~(MTM)}}
 unifies MLM and MFM in one loss function.  It is proposed by Xu et al.~\citep{Xu_VLM_ACL21} and the formula is defined as:
\begin{equation*}
 \begin{aligned}
   & \mathcal{L}_\textit{MTM} = -\mathbb{E}_{t_m \sim t} (log\text{NCE}(t_m|t_{\backslash t_m}))\\
   &  \text{NCE}(t_m|t_{\backslash t_m}) = \frac{\text{exp}(h(t_m)t_m^T)}{\mathcal{Z}} \\
   & \mathcal{Z} = \text{exp}(h(t_m)t_m^T)+\sum_{t_j \in t_{\backslash t_m}}\text{exp}(h(t_m)t_j^T)
 \end{aligned}
\end{equation*}
Compared with MLM and MFM, MTM learns joint token embeddings for both video and text tokens. Furthermore, it also expands the contrasted negative samples in two separate losses for MFM and MLM. 

\paragraph{\textbf{Masked Modal Modeling~(MMM)}}
 is first used in Univl~\\(\cite{Luo_univl_arxiv20} as part of the pre-training strategy and later is formally proposed by VLM~\citep{Xu_VLM_ACL21}. It masks either all video tokens or all text tokens, which encourages the model to use tokens from one modality to recover tokens from the other modality. The objective function employs NCE as in MTM, and experiments in VLM~\citep{Xu_VLM_ACL21} have proved its effectiveness especially for text-based video retrieval~\citep{Xu_MSRVTT_CVPR16}. 

\paragraph{\textbf{Language Reconstruction~(LR)}}
LR is a generative task, which aims to enable the pre-trained model with the ability of video caption generation. The difference between LR and masked language method~(MLM and MMM with all text tokens being masked) is that LR generates sentence from left to right, which means the model only attends to the former text tokens and video tokens when predicting the next text token.
The loss function is:
\begin{equation*}
    \mathcal{L}_{LR} = -\mathbb{E}_{w_i \sim w^{\prime}}(logP(w_i|w_{\textless i},w^{\prime},v))
\end{equation*}
where $w^{\prime}$ is the groundtruth of word sequence and $w$ is the masked version.
\paragraph{\textbf{Video Language Matching~(VLM)}}
 aims to learn the alignment between video and language. There are different task forms of VLM and we classify them into 1) Global Video Language Matching~(GVLM) and 2) Local Video Language Matching~(LVLM). 
For the GVLM, one objective function is adapted from the Next Sentence Prediction~(NSP) task used by BERT~\citep{Devlin_BERT_coRR18}, which takes in the hidden state of special token [cls] to a FC layer for binary classification. The objective function is:
\begin{equation*}
 \mathcal{L}_\textit{GVLM} = -\mathbb{E}_DlogP(y|v,w))
\end{equation*}
where y=1 if v and w are matched. 
Another VLM is to match the sequence embedding of the two modalities. Specifically, it transfers the 2 embedding sequence of video and language into 2 single feature by mean pooling or linear transfer, then it forces the paired samples closer while pushes away different ones by $\text{MIL-NCE}$~\citep{Miech_MILNCE_CVPR20} or other functions. This objective is usually used in pre-training models with multi-stream structure, which does not contain the special token [cls] for direct matching prediction. The example objective function~\citep{Luo_univl_arxiv20} is:
\begin{equation*}
    \begin{aligned}
    & \mathcal{L}_\textit{GVLM} = -\mathbb{E}_{(v,w)\sim \textbf{B}}log\text{MIL-NCE}(v,w)\\
    & \text{MIL-NCE}(v,w) = \frac{\sum_{(\widehat{v}, \widehat{w}) \in \mathcal{P}_{v,w}}(\text{exp}(\widehat{v}\widehat{w}^T))}{\mathcal{Z} } \\
    &\mathcal{Z} = \sum_{(\widehat{v}, \widehat{w}) \in \mathcal{P}_{v,w}}(\text{exp}(\widehat{v}\widehat{w}^T))+\sum_{(\widetilde{v}, \widetilde{w}) \in \mathcal{N}_{v,w}}(\text{exp}(\widetilde{v}\widetilde{w}^T))
 \end{aligned}
\end{equation*}
where $ \widehat{v} $, $\widetilde{v}$, $\widehat{w}$, $\widetilde{w}$ are mean pooling of video sequence $v$ and text sequence $w$ respectively, the negative pairs $\mathcal{N}_{v,w}$ take negative video clips or captions within the batch  $\textbf{B}$ after fixing $v$ or $w$.

Another VLM aims to align video and language locally, thus we abbreviate it as LVLM~(Local Video Language Matching). It is first proposed in HERO~\citep{Li_HERO_EMNLP20} that matches video and language at the frame level. That is, computing query-video matching score by dot product: $s=vq \in \mathbb{R}^{N^v}$, where q is the query obtained from language sequence. Two trainable 1D CNNs followed by softmax operation are applied to the matching scores to get two probability vectors $p_{st}, p_{ed}$, which represent the probability of every position being the start and the end of the ground-truth span. The objective function uses cross-entropy loss and can be summarized as:
\begin{equation*}
    \mathcal{L}_\textit{LVLM} = -\mathbb{E}_D(log(p_{st}[y_{st}])+log(p_{ed}[y_{ed}]))
\end{equation*}

\paragraph{\textbf{Sentence Ordering Modeling~(SOM)}}
SOM is first proposed in VICTOR~\citep{Lei_VICTOR_2021}, which aims to learn the relationships of text tokens from sequential perspective. Specifically, 15\% sentences are selected, randomly split into 3 segments and shuffled by a random permuted order. Therefore, it can be modeled as a 3!-class classification problem. To be specific, after multi-modal fusion, the embedding of special token [cls] is input into the FC layer followed by a softmax operation for classification. The overall objective function is:
\begin{equation*}
    \mathcal{L}_\textit{SOM} = -\mathbb{E}_D(logP(y|w_{s},v))
\end{equation*}

where y is the groundtruth of segment order and $w_s$ is the shuffled word sequence. 

\paragraph{\textbf{Frame Ordering Modeling~(FOM)}}
FOM is proposed in VICTOR~\citep{Lei_VICTOR_2021} and HERO~\citep{Li_HERO_EMNLP20}. The core idea is to  randomly shuffle a fixed percentage of frames and predict their original order.
VICTOR~\citep{Lei_VICTOR_2021} randomly selects to shuffle 15\% frames. The embedding of each shuffled frame is transformed through a FC layer, followed with softmax operation for $N_v$-class classification, where $N_v$ is the maximum length of frame sequence. 
HERO~\citep{Li_HERO_EMNLP20} also randomly selects 15\% of frames to be shuffled. The embeddings of all frames are transformed through a FC layer, followed with softmax operation to produce a probability matrix $P \in \mathbb{R}^{N_v \times N_v}$. $P_{i,j}$ represents the scores of the i-th frame that belongs to the j-th time stamp.
The two types of FOM can be summarized into one objective function:
\begin{equation*}
    \mathcal{L}_\textit{FOM} = -\mathbb{E}_D(logP(y|v_{s},w))
\end{equation*}
where y is the groundtruth of frame order and $v_{s}$ is the shuffled frame sequence. 

\begin{table*}[t]
    \centering
    \begin{tabular}{c|c|c|c|c}
    \toprule
        Task &  Type&Strategy  &Sub-task & Description  \\
        \hline
        
         MLM& Completion & USL &~ &  Predicting text tokens that are masked with certain percentage.
        \\
         \hline
         \multirow{3}{*}{MFM}& \multirow{3}{*}{Completion} &  \multirow{3}{*}{USL} &  MFMCE &Predicting masked frame tokens with cross entropy loss.
          \\
          \cline{4-5}
         ~ &  ~ &  ~ &MFMR & Reconstructing the masked video tokens with regression loss.
         \\ 
          \cline{4-5}
         ~ &  ~ &  ~ & MFMCL &\makecell[c]{\textcolor{black}{Identifying the masked video tokens from negative samples} \\ constructed by various methods.} 
         \\
         \hline
         MTM & Completion &  USL &~ & \makecell[c]{\textcolor{black}{Identifying the masked tokens~(video or text) from negative samples} \\ constructed by various methods.}
         \\
         \hline
         MMM & Completion &  USL&~ & \makecell[c]{Masking either all video tokens or all text tokens  and recovering them \\from the other modality.}
         \\
         \hline
         LR & Completion &  SL &~ &  Generate text sequence from left to right according to video modality.
          \\
         \hline
         \multirow{2}{*}{VLM} &  \multirow{2}{*}{Matching} &  \multirow{2}{*}{USL}  &GVLM & Globally matching video and text modality.
         \\
         \cline{4-5}
         ~ &~& ~ & LVLM & Matching video and text at the frame level.
         \\
         \hline
         SOM &  Ordering  & USL&~& \makecell[c]{Randomly shuffling sentence and reconstruct the sentence order \\from video modality.} 
         \\
         \hline
         FOM & Ordering & USL&  ~&\makecell[c]{Randomly shuffling video tokens and reconstruct the frame order \\from text modality.}
          \\
          
    \bottomrule
    \end{tabular}
    \caption{A summary of proxy tasks that commonly used in Video-Language Pre-training.}
    \label{tab:proxy_tasks}
\end{table*}

\subsubsection{Video-Language Downstream Tasks}
The target of pre-training is to better adapt the learned knowledge from a large corpus to downstream tasks via transfer learning~\citep{Belinkov_TL_ACL17}.
Representative downstream tasks also play the role in evaluating pre-trained models.
For better transfer impact, we need to consider the model structure and choose appropriate transferring method for each downstream task.
The common downstream tasks that appear in the Video-Language pre-training include generative tasks and classification tasks. We introduce the task requirements and how to transfer the knowledge from pre-training to downstream tasks in the following subsections.

\paragraph{\textbf{Text-based Video Retrieval}}
~\citep{Yu_VideoRetrieval_ECCV18} is defined to retrieve a
relevant video/video segment given an input text query.
It requires model to map the video and text modality into a common semantic embedding space. Since the proxy task of VLM aims at learning the alignment between video and  text, many works~\citep{Zhu_ActBERT_CVPR20,Li_HERO_EMNLP20,Luo_univl_arxiv20,Lei_ClipBert_CVPR21} adapt the proxy task of VLM to calculate the matching score of these two modalities directly.  

\paragraph{\textbf{Action Recognition}}
~\citep{Zhu_ActionRecog_arxiv20} is defined to classify the action category of the given video/video segment, which is a representative classification task for video understanding. To transfer pre-trained knowledge to action recognition, works in~\citep{Sun_CBT_ECCV20,Lei_VICTOR_2021} use the pre-trained models as feature extractors and finetune a linear classifier added on the top of pre-trained model for action recognition.     

\paragraph{\textbf{Action Segmentation}}
~\citep{Ding_ActionSegmentation_CVPR18} is designed to predict action label of given video/video segment at the frame level. It is also a classification task with video as the only input. To apply pre-trained models to action segmentation, several works~\citep{Zhu_ActBERT_CVPR20,Xu_VLM_ACL21} use the pre-trained models as feature extractors and add a linear classifier upon the extracted video features. 

\paragraph{\textbf{Action Step Localization}}
 is first proposed in Cross Task \citep{Zhukov_CrossTask_CVPR19}, which aims to recognize action steps in instructional videos. The difference between action step localization and action recognition is that for the step localization, event is described with manual phrase but not from fixed label dictionary. To apply pre-trained models to action step localization, works in~\citep{Zhu_ActBERT_CVPR20,Luo_univl_arxiv20,Xu_VLM_ACL21}  regard manual phrase as text description and calculate its relevance score with input video/video segment by either dot production or linearly transforming the embedding of [cls] . 

\paragraph{\textbf{Video Question Answering}}
~\citep{Tapaswi_MovieQa_CVPR16,Lei_TVqa_EMNLP18,Jang_TGIFqa_CVPR17} aims to automatically  answer natural language questions given a context video. VideoQA applied in Video-Language pre-training can be divided into multiple choices task or fill-in-the-blank task according to the types of the answers, both of which can be handled as classification tasks. 
For multi-choice VideoQA, works in~\citep{Zhu_ActBERT_CVPR20,Li_HERO_EMNLP20} feed candidate answer at the end of query sentence to generate QA-aware global representations, and input the global representations into MLP based classifier to obtain the matching score. The final choice is made by selecting the candidates with the max matching score. For fill-in-the-blank VideoQA, ActBERT~\citep{Zhu_ActBERT_CVPR20} proposes a similar method, which adds a linear classifier upon the cross-modal feature but without the input of candidate text.

\paragraph{\textbf{Video Captioning}}
~\citep{Chen_videoCaption_IJCAI19, Zhou_VideoCaption_CVPR18} is the task of generating a natural-language utterance for the given video, which is the only generative task among the downstream tasks introduced in this paper. It is one of the most typical tasks for multi-modal understanding and nearly all works related to Video-Language pre-training evaluate their pre-trained models on this task. To transfer pre-trained knowledge to video captioning, works in~\citep{Sun_videoBERT_ICCV19, Zhu_ActBERT_CVPR20, Li_HERO_EMNLP20} use pre-trained models as video feature extractor or video encoder and  add a transformer-based decoder for finetuning. 
Works in~\citep{Xu_VLM_ACL21} transfer a single encoder to generate word sequence by reusing the pre-trained model as prediction heads.
Work in~\citep{Luo_univl_arxiv20} includes a generative task in the pre-training stage by adding a transformer decoder, which reduces the gap between the proxy task and the video captioning task.

As shown in above introduction, Video-Language pre-training works focus more on classification task. Improving the pre-trained model's ability especially for generation can be further explored.
What's more, in addition to the downstream tasks we listed above, other downstream tasks such as multi-modal sentiment analysis~\citep{Zadeh_MultimodalAnalysis_EMNLP17}, image-based retrieval~\citep{Mao_ImageBasedRetrieval_IEEE17} have also been explored recently.

\subsection{Video-Language Datasets}
\label{section:datasets}
Compared with CNNs, transformer based frameworks rely heavily on massive datasets especially for pre-training. The quality and quantity of video datasets matter a lot to model's performance. In this section, we divide the commonly used video datasets into 3 categories according to the types of their annotations: label-based datasets, caption based datasets andother datasets. 
Tab.~\ref{tab:datasets} summarizes all mentioned datasets.

\begin{table*}[t]
    \centering
    \begin{tabular}{lcccccc}
    \toprule
        Dataset &  videos &clips & annotations &  duration & source & year  \\
        \hline
        \textbf{Label Based Datasets} \\
        
         HMDB51 & 3.3k & 6.8k &  labels &24h& Web/Other Datasets & 2011 
         \\ 
         \hline
         
         UCF101 & 2.5k & 13.3k & labels & 27h & YouTube & 2012
         \\
         \hline
        
         MPII Cooking& 44 & 5.6k &labels & 8h& Kitchen & 2012
         \\
         \hline
         Kinetics400 & 306k & 306k & labels & 817h & YouTube &2017
         \\
         \hline
        AVA &430 & 230k & labels & 717h & YouTube & 2018
        \\
        \hline
         \hline
         
         \textbf{Caption Based Datasets} \\
          Howto100M & 1.22M & 136M & 136M captions &134,472h & YouTube & 2019 
        \\
        
         \hline
         Alivol-10M* & 10.3M & 11M & 11M captions & 98,801h & e-commerce  &2020
          \\
          \hline
        Auto-captions on GIF & 163k & 163k & 164k & - &GIF Web & 2020 \\
          
          \hline
         ActivityNet &20k & 100k &100k captions & 849h&YouTube&2015 
         \\
         
         \hline
         Charades & 10k & 18k & 16k captions& 82h & Home & 2016 
         \\
         \hline
         TGIF&102k  & 102k & 126k captions & 103h & Tumblr GIFs & 2016
         \\
         \hline
         YouCookII& 2k & 14k & 14k captions &176h &YouTube &2016 
         \\
         \hline
         MSR-VTT & 7.2k & 10k & 200k captions & 40h & YouTube & 2016
         \\
         \hline
         Didemo & 10k& 27k &41k captions &87h &Fliker&2017
         \\
         \hline
         LSMDC& 200 & 128k &128k captions& 150h& Movies&2017 
         \\
         \hline
         
         How2& 13k & 185k &185k captions & 298h & YouTube & 2018
         \\

         \hline
         TVR& 21.8k & 21.8K & 109k queries & 460h & TV shows & 2020
         \\
         \hline
         TVC & 21.8k &21.8k  & 262k captions& 460h & TV shows & 2020
         \\
         \hline
         
         VIOLIN&  6.7k &16k   & 95k  captions& 582h & Movie \& TV show & 2020 
         \\
         \hline
         \hline
         \textbf{Other Datasets} \\
         TVQA & 925 & 21.8k & 152.5k QAs &460h & TV shows & 2018 
         \\
         \hline
         COIN&12k  &  46k & segment labels  & 476h & YouTube& 2019
         \\
         \hline
         CrossTask& 4.7k & 20k & 20k steps & 376h & YouTube& 2019
         \\
         \hline
         
    \bottomrule
    \end{tabular}
    \caption{Commonly used Datasets in Video-Language Pre-training and finetuning. suffix * for Alivol-10 means that the dataset is not released yet. We divide the datasets into 3 groups according to the type of their annotations: Label Based Datasets, Caption Based Datasets and Other Datasets.}
    \label{tab:datasets}
\end{table*}

\subsubsection{Label Based Datasets}
Label Based Datasets are the datasets with labels at the video level. They are widely used for classification tasks such as action recognization.
For example, HMDB51~\citep{Kuehne_HMDB_ICCV11} contains 6,841 videos from 51 action categories in total. 
UCF101~\citep{Soomro_UCF101_arxiv12}, MPII Cooking~\citep{Rohrbach_MPIICooking_CVPR12}, Kinetics series~\citep{Kai_Kineics400_arxiv17} and  AVA~\citep{Gu_AVA_CVPR18} are the other representative datasets. 

\subsubsection{Caption Based Datasets}
Caption Based Datasets require descriptions for each video or video segment. For example, Activitynet~\citep{Krishna_anet_ICCV17} includes 20k YouTube untrimmed videos with 100k manually caption sentences. Each caption describes the content of the corresponding video segment annotated by start and end time stamps. 
Caption is the major annotation of video datasets with widely applications. 
On the one hand, large-scale Caption Based Datasets can be used for Video-Language pre-training. For instance, Howto100M~\citep{Miech_Howto100m_ICCV19} is so far the largest English video dataset, which contains 136M video clips paired with captions from YouTube (mostly instructional videos), most works~\citep{Sun_CBT_ECCV20,Zhu_ActBERT_CVPR20,Li_HERO_EMNLP20} pre-train their models on this dataset.
Alivol-10M~\citep{Lei_VICTOR_2021} is a Chinese e-commerce dataset with 10M videos of 98,801 hours in total. The descriptions mostly follow the
standards of the e-commerce platform to describe the visual content of certain product.
Auto-captions on GIF~\citep{Pan_GIFs_arxiv20} is newly designed for generic video understanding based on GIF videos. The paired description is extracted from the
Alt-text HTML attribute of each GIF video. 
On the other hand, datasets with caption annotations are widely used in downstream tasks such as video retrieval/video moment retrieval, video captioning/dense video captioning and text based localization~(requires time stamps annotations). As shown in Tab.~\ref{tab:datasets},  ActivityNet~\citep{Krishna_anet_ICCV17}, Charades~\citep{Sigurdsson_CHarades_ECCV16}, TGIFs~\citep{Li_TGIF_CVPR16}, YouCookII~\citep{Zhou_youcook2_AAAI18}, etc. are the representative caption datasets.

\subsubsection{Other Datasets}
In addition to the caption and label annotations, other types of annotations are used for other downstream-tasks. As shown in Tab.~\ref{tab:datasets},
TVQA~\citep{Lei_TVqa_EMNLP18} is a videoQA dataset based on 6 popular TV shows, with 460 hours of videos and 152.5K human-written QA pairs in total. Each query provides 5 candidates with one correct answer, the correct answer is also marked with start and end time stamps for further inference.
COIN~\citep{Tang_COIN_CVPR19} is designed for \textbf{CO}mprehensive \textbf{IN}structional video analysis, which is organized with a 3-hierarchical structure, from domain, task, to step.
The dataset contains 11,827 instructional videos in total with 12 domains, 180 tasks, and 778 pre-defined steps.
As all the videos are annotated  with a series of step descriptions and the corresponding temporal boundaries, COIN is commonly used for  action segmentation task.
CrossTask~\citep{Zhukov_CrossTask_CVPR19} contains 4.7k instructional videos crawled from YouTube, related to 83 tasks. For each task, an ordered
list of steps with short descriptions are provided. 
Works in~\citep{Zhu_ActBERT_CVPR20, Luo_univl_arxiv20} evaluate their pre-trained models on the task of Action Step Localization~\citep{Zhukov_CrossTask_CVPR19} based on this dataset.

\begin{figure*}[t]
  \centering
  \includegraphics[width=\textwidth]{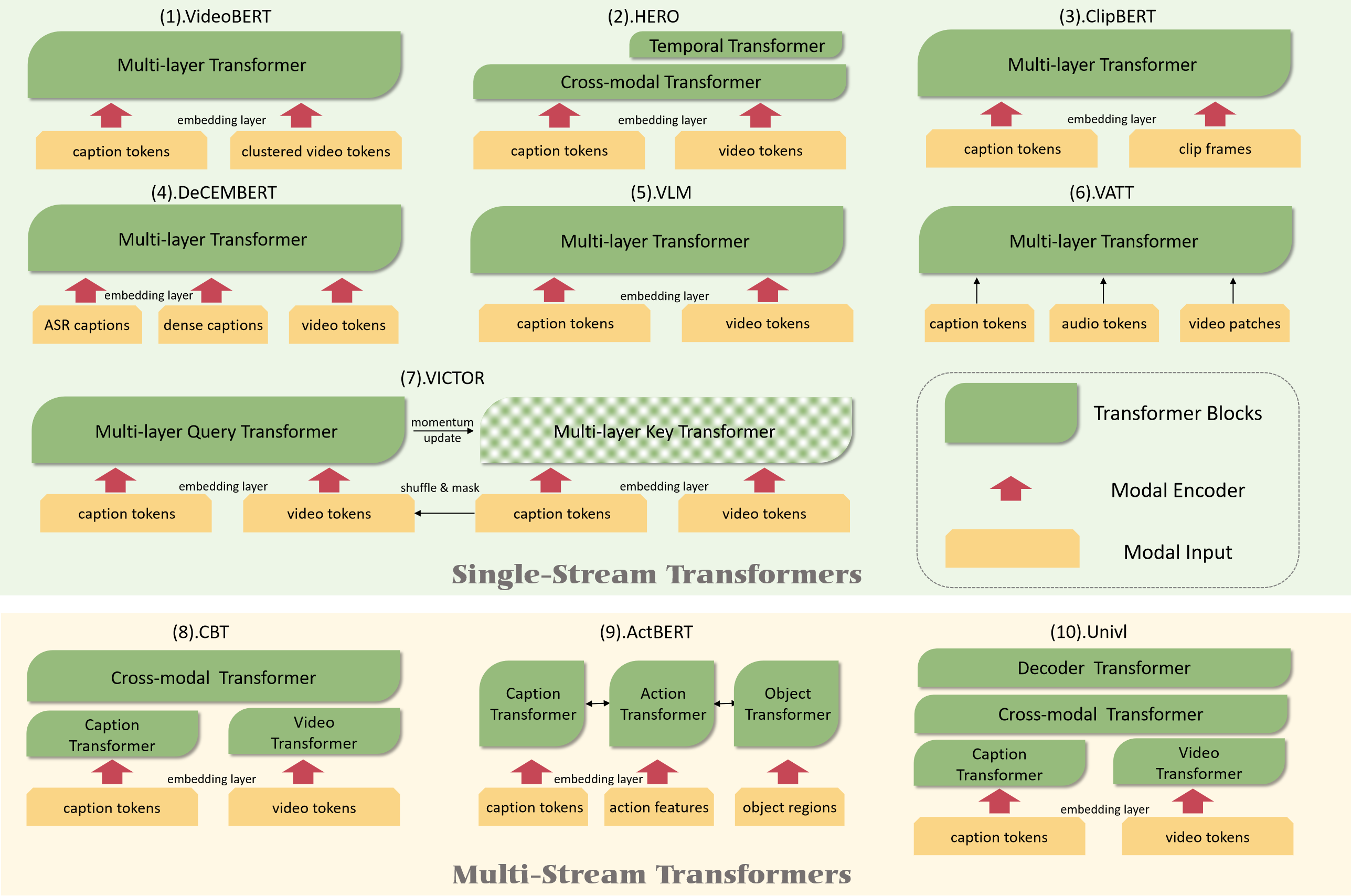}
  \caption{An overview of Transformer models used for Video-Language representation learning. All models are divided into \textbf{Single-Stream Transformers}~(VideoBERT~\citep{Sun_videoBERT_ICCV19}, HERO~\citep{Li_HERO_EMNLP20}, ClipBERT~\citep{Lei_ClipBert_CVPR21}, DeCEMBERT~\citep{Lei_ClipBert_CVPR21}, VLM~\citep{Xu_VLM_ACL21}, VATT~\citep{Akbari_VATT_arxiv21}, VICTOR~\citep{Lei_VICTOR_2021} ) and \textbf{Multi-Stream Transformers}~(CBT~\citep{Sun_CBT_ECCV20}, ActBERT~\citep{Zhu_ActBERT_CVPR20}, Univl~\citep{Luo_univl_arxiv20}) according to their structure. Despite the differences in model structure, most models take caption tokens and video tokens as inputs, while DeCEMBERT takes ASR captions as additional text information, ActBERT takes object regions as  additional visual information and VATT takes audio as additional modality information. As for modal encoders, most models apply modality encoders to extract modality features while VATT abandons them. }
  \label{fig:transformer_models}
\end{figure*}

\begin{table*}[t]
    \centering
    \begin{tabular}{llllll}
    \toprule
        Method & Inputs & Proxy Tasks & Pre-train Dataset &  Downstream Task(dataset) & Source  \\
        \hline
        \hline
        \multicolumn{2}{l}{\textbf{Single-Stream}}  \\
        VideoBERT & video+text & \makecell[l]{MLM, MFM,\\VLM} & Cooking & \makecell[l]{Action Classification(YoucookII),\\ Video Caption(YoucookII)}  & ICCV  \\
        \hline
        
        HERO & video+text & \makecell[l]{MLM, MFM, \\VLM, FOM} &\makecell[l]{Howto100M,\\TV} &\makecell[l]{Video retrieval(MSR-VTT,\\TVR,DiDeMo,How2R), \\
        VideoQA(TVQA,How2QA), \\
        Video-and-Language \\Inference (VIOLIN), \\
        Video Caption(TVC)} & EMNLP\\
        \hline
        
        ClipBERT & clip+text & MLM, VLM & \makecell[l]{COCO Captions,\\VG Captions} & \makecell[l]{Video 
        Retrieval(MSR-VTT,\\ DiDeMo,ActivityNet),\\
        VideoQA(TGIF-QA,MSRVTT)} & CVPR\\
        \hline
        
        DeCEMBERT & video+text  &\makecell[l]{MLM, VLM,\\Constrained \\Attention Loss} & Howto100M& \makecell[l]{Video Retrieval(MSR-VTT, YouCookII),\\ Video Caption(MSR-VTT, YouCookII), \\
        VideoQA(MSRVTT)} &NAACL \\
        \hline
        
        VLM & video+text &\makecell[l]{MTM, MMM} & Howto100M & \makecell[l]{Video retrieval(MSR-VTT, YouCookII),\\ Action Segmentation(COIN),\\
Action Step Localization(CrossTask),\\
VideoQA(MSR-VTT),\\
Video Caption(YouCookII)} & ACL\\
        \hline
        
        VATT & \makecell[l]{video+text\\+audio} & \makecell[l]{Multi-modal \\Contrastive\\ Learning} & \makecell[l]{Howto100M,\\
AudioSet} &  \makecell[l]{Action Recognition(UCF101,\\HMDB51,kenitics-400,600),\\
Audio Event Classification\\(ESC50,AudioSet), \\Video Retrieval(MSR-VTT,YouCookII),\\Image Classification(ImageNet)} & arxiv\\
        \hline
        
        VICTOR & video+text & \makecell[l]{MLM, MFOM, \\MSOM, dual-VSA,\\ intra-MFM, \\inter-MFM} & Alivol-10M & \makecell[l]{Video Retrieval(Alivol-10M), \\Video Classification(Alivol-10M),\\ Video Recommendation\\(users'
video viewing logs),\\Video Caption(Alivol-10M)} & arxiv\\
        \hline
        \hline
        \multicolumn{2}{l}{\textbf{Multi-Stream }}  \\
        CBT &video+text &\makecell[l]{MLM, MFM,\\VLM} &  Howto100M &\makecell[l]{Action Recognition(UCF101,\\HMDB51), \\
Action Anticipation(Breakfast,\\50Salads, ActivityNet),\\ Video Caption(YouCookII),\\ Action Segmentation(COIN)} & ECCV\\
        \hline
        
         ActBERT & \makecell[
         c]{action+region\\+text} & \makecell[l]{MLM, MAM,\\
MOM, VLM} & Howto100M & \makecell[l]{Video Retrieval(YouCookII,\\ MSR-VTT),\\Video Caption(YouCookII),\\Action Segmentation(COIN),\\Action Step Localization(CrossTask),\\
VideoQA(LSMDC, MSR-VTT)} & CVPR \\
         \hline
         
         Univl & video+text &  \makecell[l]{MFM, MLM,\\ VLM, LR } & Howto100M & \makecell[l]{Video Retrieval(YouCookII, \\MSR-VTT),\\
Video Caption(YoucookII),\\
Action Segmentation(COIN),\\
Action Step Localization(CrossTask),\\
Multi-modal Sentiment Analysis\\(CMU-MOSI)} & arxiv\\
         
    \bottomrule
    \end{tabular}
    \caption{A summary of Video-Language Pre-training methods.}
    \label{tab:video_language_pretraining}
\end{table*}

\section{Video-Language Transformer Models}
In this section, we provide an overview of Transformer based models for Video-Language pre-training in Fig.~\ref{fig:transformer_models}. 
We roughly divide different models into two categories based on their model structure: Single-Stream Transformers and Multi-Stream Transformers. 
For the Single-Stream Transformers,  features/embeddings of different modalities are input into a single transformer to capture their intra and inter modality information. Multi- Stream Transformers input each modality into independent transformers to capture information within modalities and then build cross-modal relationship via for example another transformer. In addition to the model structure, the distinctions across different methods relate to their inputs, proxy tasks and downstream tasks and benchmarks, which we summarize in Tab.~\ref{tab:video_language_pretraining} and describe in details below.

\subsection{Single-Stream Structure}
\subsubsection{VideoBERT}
VideoBERT~\citep{Sun_videoBERT_ICCV19} is the first to explore Video-Language representation with transformer based pre-training method. It follows the single-stream structure, porting the original BERT structure to the multi-modal domain as illustrated in Fig. ~\ref{fig:transformer_models}-(1). Specifically, it inputs the combination of video tokens and linguistic sentence into multi-layer transformers, training the model to learn the correlation between video and text by predicting masked tokens. VideoBERT shows the ability of simple transformer structure  to learn high level video representations that capture semantic meaning and  long-range temporal dependencies.

To discretize continuous videos as discrete word tokens, they cut the video into small clips of fixed length and cluster the tokens to build a video dictionary.
In pre-training stage, the model is trained with proxy tasks of MLM, MFM and VLM, corresponding to the feature learning in text-only domain, video-only domain, and video-text domain. Although with the simple proxy tasks and plain model structure, VideoBERT shows great performance on the downstream tasks of zero-shot action classification and video captioning. 
The model is initialized with the pre-trained BERT weights, the video token is generated based on the S3D~\citep{Xie_S3D_ECCV18} backbone.  All experiments are applied on the cooking domain, with pre-training on the large scale of cooking videos crawled from YouTube by authors themselves and evaluating on the YouCookII benchmark dataset~\citep{Zhou_youcook2_AAAI18}.

\subsubsection{HERO}
As illustrated in Fig.~\ref{fig:transformer_models}-(2), Li et al.~\citep{Li_HERO_EMNLP20} propose HERO, a \textbf{H}ierarchical \textbf{E}ncode\textbf{R} for \textbf{O}mni representation learning, which contains a cross-modal transformer to fuse video frame sequence and corresponding sentence, and a temporal transformer to learn contextualized video embeddings from the global context.
Previous works simply adapt proxy tasks of masking~(MLM) and matching~(VLM) that originated from NLP domain. HERO firstly designs the proxy tasks of LVLM~(Local Video Language Matching) and FOM~(Frame Order Modeling), which consider the sequential nature of videos. These two proxy tasks have been described in Section~\ref{section:proxy_tasks}. The experiments of HERO prove that hierarchical transformer structure and new proxy tasks are both beneficial to downstream tasks.
Li et al.~\citep{Li_HERO_EMNLP20} also expand the pre-training datasets from instructional video domain to TV or movie domain. They find that text-based
video-moment retrieval is more sensitive to
domain gaps. In other words, keeping dataset domain consistent, text-based video retrieval could achieve the same or better performance with less pre-training data.  

To be more specific, HERO extracts both 2D and 3D video features with ResNet~\citep{He_ResNet_CVPR16} and  Slow-Fast~\citep{Feichtenhofer_SlowFast_ICCV19} respectively. 
The cross-modal transformer takes the combination of video sequence and text sequence as input to learn contextualized embeddings through cross-modal attention. 
The output of visual embeddings are further input into temporal transformer to learn  contextualized embeddings from the global video context.
HERO applies the proxy tasks of MLM, MFM, VLM and FOM in pre-training stage and transfers to downstream tasks of video retrieval, videoQA, video-and-language inference and video captioning. 
The ablation study shows that FOM can effectively benefit downstream tasks that rely on temporal reasoning (such as QA tasks), VLM for both global and local alignment can benefit the retrieval tasks.

\subsubsection{ClipBERT}
Lei et al.~\citep{Lei_ClipBert_CVPR21} propose a generic framework ClipBERT for video-text representation learning that could be trained in end-to-end manner. 
Different from previous works that extract video features from pre-trained backbone such as S3D~\citep{Xie_S3D_ECCV18}, ClipBERT directly samples a few frames from each video clip, using 2D CNN as backbone instead of 3D CNN for lower memory cost and better computation efficiency. 
Based on 2D visual backbone, they also demonstrate that image-text pre-training on COCO~\citep{Chen_CoCo_Arxiv15} and Visual Genome Captions~\citep{Krishna_VG_IJCV17} benefits video-text tasks. 
ClipBERT adopts a sparse sampling strategy, including sampling a few frames from each clip and using only a single or a few sampled clips instead of full-length videos. The experiments show 1 or 2 frames per clip and 2 clips per video is sufficient for effective Video-Language pre-training.

The concrete structure of ClipBERT is single-stream \ (Fig.~\ref{fig:transformer_models}-(3)), the video input is patch sequence of a single clip.
After 2D backbone generates T visual feature map for T frames of each single clip, a temporal fusion layer is applied to aggregate the frame-level feature maps into a single clip-level feature map. 
A cross transformer is then applied to combine the clip feature map and text sequence to capture the cross-modal relationship. During the inference, when multiple clips are used, the predictions are fused together as the final output. ClipBERT uses MLM and VLM objectives to optimize the model, the pre-trained weights are further finetuned to text-based video retrieval and videoQA on 6 benchmarks.

\subsubsection{DeCEMBERT}
Tang et al.~\citep{Tang_DeCEMBERT_NAACL21} propose the approach of \textbf{De}nse \textbf{C}aptions and \textbf{E}ntropy \textbf{M}inimization~(DeCEMBERT) to alleviate the problem that the automatically generated captions in pre-training dataset like Howto100M~\citep{Miech_Howto100m_ICCV19} are noisy and sometimes unaligned with video content.
To be specific, the original caption may not describe the rich content of the corresponding video or contains only irrelevant words due to recognition error of ASR. Therefore, DeCEMBERT uses dense captions~\citep{Johnson_DensCap_CVPR16} generated from ~\citep{Yang_DenseCap_CVPR17} as additional language input for the model learning.
To better align video with ASR captions, DeCEMBERT  propose a constrained attention loss that encourages the model to select the best matched ASR caption from a pool of continuous caption candidates. 

As illustrated in Fig.~\ref{fig:transformer_models}-(4), DeCEMBERT applies the single-stream structure, using a BERT like transformer to  encode the relationship of video features, dense captions and a set of continuous ASR captions. 
The whole model is pre-trained with MLM, VLM tasks and finetuned on video captioning, text-based video retrieval and videoQA.
Comprehensive experiments demonstrate that DECEMBERT is an improved pre-training method for learning from noisy, unlabeled dataset.

\subsubsection{VLM}
Previous methods~\citep{Luo_univl_arxiv20,Li_HERO_EMNLP20} propose either multiple transformer encoders or a single cross-modal encoder but requires both modalities as inputs, What's more, existing pre-training tasks tend to be more and more task-specific, limiting the extensibility and generalization ability of pre-trained models. In contrast, VLM~(\textbf{V}ideo-\textbf{L}anguage
\textbf{M}odel) is a task-agnostic model with BERT like cross-model Transformer that can accept text, video, or both as input.

VLM introduces two new schemes of masked tasks: Masked Modality Modeling~(MMM) and Masked Token Modeling~(MTM). MMM is to randomly mask a whole modality for a portion of training examples, which forces the encoder  to reconstruct the masked modality based on the tokens from the other modality. 
MTM is to randomly mask a fixed portion of tokens (both video or language tokens) and predict them from negative candidates, which unifies the losses on MLM and MVM.
MMM has been validated to be effective especially for text-based video retrieval and MTM performs better than MLM+MVM. 
VLM is evaluated on the downstream tasks of text-based video retrieval, action segmentation, action step localization and videoQA. To apply the BERT like model with single encoder to generative tasks such as video captioning, VLM uses a masked attention map to make the future text tokens unavailable. Based on that, VLM re-use the
language model heads as prediction heads for generation with no extra decoder architecture. 
Experimental results show that VLM can maintain competitive performance while requiring less parameters.

\subsubsection{VATT}
Akbari et al.~\citep{Akbari_VATT_arxiv21} present an end-to-end framework VATT~(\textbf{V}ideo-\textbf{A}udio-\textbf{T}ext \textbf{T}ransformer) for leaning multi-modal representations from raw video, audio and text.
To be specific, they partition the raw video frames into a sequence of $[T/t]\times[H/h]\times[W/w]$ patches, where T, H, W correspond to video's temporal, height, width dimension respectively. The raw audio waveform is segmented on its temporal dimension. The word token is represented by one-hot vector. These three modality sequences are transformed by linear projection but not pre-trained backbones as previous works do.
To obtain inherent co-occurrence relationships of three modalities, Akbari et al.~\citep{Akbari_VATT_arxiv21} adopt the most widely used transformer  architecture~(ViT)  except keeping the layer of tokenization
and linear projection reserved for each modality separately. 
VATT is optimized by matching video-audio pairs and video-text pairs with common space projection and contrastive learning. 
The whole model is pre-trained on Howto100M~\citep{Miech_Howto100m_ICCV19} providing video-audio-text triplets  and AudioSet~\citep{Gemmeke_AudioSet_ICASSP17} providing audio-text pairs. After pre-training, VATT is finetuned on the downstream tasks of action recognition, audio event classification~\citep{Dai_AudioEvent_ICASSP17}, text-based video retrieval and image classification. The experiment results of image classification on ImageNet~\citep{Deng_Imagenet_2009} demonstrate that VATT can be adapted from video domain to image domain.

In conclusion, VATT validates that large-scale self
supervised pre-training is a promising direction to learn multi-modal representation~(video, text, audio) with pure attention-based model and end-to-end training.

\subsubsection{VICTOR}
VICTOR~\citep{Lei_VICTOR_2021} stands for  \textbf{VI}deo-language
understanding via \textbf{C}ontrastive mul\textbf{T} im\textbf{O}dal p\textbf{R}e-training, which is trained on Chinese Video-Language dataset. 
VICTOR follows the single-stream model structure, with an encoder transformer to obtain the cross-modal relationship, a decoder transformer for generative tasks. What's more, inspired by MoCo~\citep{He_MoCo_CVPR20} that expands negative samples with memory bank and momentum updating for better contrastive learning, VICTOR involves memory queues that save the negative samples for calculating contrastive losses. Synchronously, another network symmetric to the main Query network named Key network is applied to embed negative samples.

Due to the absence of Chinese pre-training dataset, Lei et al.~\citep{Lei_VICTOR_2021} collect Alivol-10M from e-commerce platform with standard descriptions and corresponding product videos. The details have been described in Section \ref{section:datasets}.
 Lei et al.~\citep{Lei_VICTOR_2021} design new proxy tasks of Masked Frame Order Modeling~(MFOM), Masked Sentence Order Modeling~(MSOM) and Dual Video and Sentence Alignment~(dual-VSA) for pre-training, where MFOM is to explore the sequential structure of videos by reordering the shuffled video sequence, MSOM is similar to MFOM but from the text perspective. For the dual-VSA~(similar with VLM), they only take matched video-text pairs as inputs, utilizing the representation of frames/text to retrieve the representation of corresponding text/frames. In other words, the negative samples come from memory bank  would only go through Key transformer network as the authors point out that inputting in the mismatched video and text would hamper the pre-training process of multi-modal encoder.
 The pre-trained weights of VICTOR are further transferred to downstream tasks of  multi-level video classification, content-based video recommendation, multi-modal video captioning, and cross-modal retrieval that with both text and image as input query. 

\subsection{Multi-Stream Structure}
\subsubsection{CBT}
CBT~\citep{Sun_CBT_ECCV20} propose noise contrastive estimation~(NCE)~\citep{Jozefowicz_NCEloss_arxiv16} as the loss objective for Video-Language learning, which preserves the fine-grained information of video compared to vector quantization~(VQ) and softmax loss in VideoBERT. 
The model contains 3 components as shown in Fig.~\ref{fig:transformer_models}-(8): one text transformer~(BERT) to embed discrete text features, one visual transformer that takes in the continuous video features and a third cross-modal transformer to embed mutual information between two modalities. CBT extends the BERT structure to multi-stream structure and verifies the effectiveness of NCE loss for learning cross-modal features.

In pre-training stage, two single modal transformers learn video and text representations respectively via contrastive learning. The third cross-modal transformer combines the two modal sequences, computes their similarity score and learns the relationship of paired video and sentence by NCE loss. Sun et al.~\citep{Sun_CBT_ECCV20} propose curriculum learning strategy by first pre-training the S3D~\citep{Xie_S3D_ECCV18} backbone and then finetuning the last block of S3D with visual transformer using visual loss. 
Both pre-trained visual features and cross-modal features are evaluated on downstream tasks of action recognition, action anticipation, video captioning and video segmentation. 

\begin{figure}[t]
  \centering
  \includegraphics[width=0.9\linewidth]{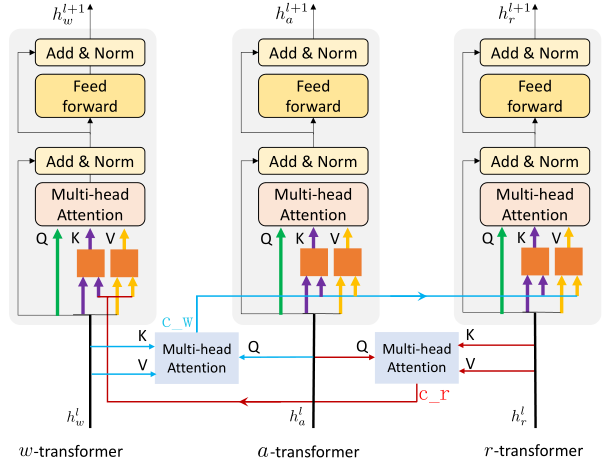}
  \caption{Illustration of Tangled Transformer block. Figure is from~\cite{Zhu_ActBERT_CVPR20}.}
  \label{fig:tangled_transformer}
\end{figure}

\subsubsection{ActBERT}
Zhu et al.~\citep{Zhu_ActBERT_CVPR20} introduce global action and local regional objects as visual inputs to learn joint video-text representations. ActBERT is a multi-stream model~(Fig.~\ref{fig:transformer_models}-(9)) with Tangled Transformer block to enhance the communications between different sources, which is illustrated in Fig.~\ref{fig:tangled_transformer}. Previous multi-stream structure always use an extra transformer layer to encode inter relationship of multi-modal information, while the Tangled Transformer block uses co-attentional transformer layer~\citep{Lu_ViLBERT_NeurIPS19} that the key-value pairs from one modality could pass through the other modality. Experiments on various Video-Language related downstream tasks verify that the global action information and local object clues are complementary.

For the global action input, they extract verbs from the corresponding descriptions of each video clip and build a verb dictionary. Then a 3D network classifier is trained to predict the each video clip's verb labels. The action feature of each clip is extracted from the 3D network classifier after global averaging layer.  
For the input of local object regions, authors use pre-trained  Faster-RCNN~\citep{Ren_FasterRcnn_NeurIPS15} to extract the  bounding boxes and the corresponding visual features. 
ActBERT is pre-trained on the proxy tasks of MLM, MAM~(Masked Action Modeling), MOM~(Masked Object Modeling) and VLM. The pre-trained weights are further transferred to 5 downstream tasks of video captioning, action segmentation, action step localization, video retrieval and videoQA. 

\subsubsection{Univl}
Previous multi-modal models are pre-trained on understanding tasks, which leads to  discrepancy for generative downstream tasks such as video captioning. Univl~\citep{Luo_univl_arxiv20} is the first one to pre-train model on both understanding and generative proxy tasks.
Univl follows multi-stream structure as illustrated in Fig.~\ref{fig:transformer_models}-(10), which contains two single transformer encoders to embed video and text respectively, a cross-modal transformer to fully interact the text and video embeddings, a decoder for generation tasks. 

Univl uses VLM, MFM, MLM and LR~(Language Reconstruction) as proxy tasks for pre-training, transfers to the downstream tasks of  text-based video retrieval, multi-modal video captioning, action segmentation, action step localization and multi-modal sentiment analysis.
There are two types of VLM in Univl. The first one is to train two single modal encoders by matching their video and text sequence with NCE loss. The other is to train the cross modal transformer by inputting the special token [cls] to predict the alignment score of given video and sentence.
The experiments show that the later type of VLM applied on the cross-modal transformer benefits more on retrieval tasks.
Univl develops a three stage training strategy for pre-training. 
Firstly, Univl trains the weights of text BERT and video transformer by matching their output sequences with NCE objective. 
Next, the whole model is trained by all objectives with smaller learning rate. 
Furthermore, Univl enhances its video representations by masking the whole text tokens with a 15\% possibility. 
The step-by-step training strategy improves the pre-training process consistently.

\subsection{Summary \& Comparison }
In this part, we provide a summary and comparison of the above mentioned methods from the perspectives of model structure, proxy tasks, training strategy,  and performance on widely used benchmarks: MSR-VTT for text-based video retrieval and YouCookII for video captioning.
The paradigm of the above methods can be summarized as building models containing transformer encoders to learn intra and inter modality representations, pre-training models on pre-designed proxy tasks and finetuning/evaluating on varies downstream tasks. 
\paragraph{\textbf{Model Structure}}
For the transformer blocks, most works apply the original transformer structure directly, while some works make adjustments to adapt to multi-modal processing. For example, VATT shares the weight of self attention layer  across different modalities, but keeps the layer of tokenization
and linear projection independent for each modality. VLM uses different attention masks to accommodate downstream tasks that require different modalities.  ActBERT use Tangled Transformer blocks to build relationship of different modalities across independent transformer blocks. 

For the word embedding, most works apply WordPiece embeddings~\citep{WuSCLNMKCGMKSJL_wordPiece_arxiv16} with a 30,000 token vocabulary provided by BERT~\citep{Devlin_BERT_coRR18}. For the video embedding, most works extract video features with fixed visual backbone S3D~\citep{Xie_S3D_ECCV18}, which is pre-trained by Miech~\citep{Miech_MILNCE_CVPR20}. There exists exceptions, for example,  VICTOR~\citep{Lei_VICTOR_2021}  utilizes 2D backbone of Inception-V4~\citep{Szegedy_Inceptionv4_AAAI17} pre-trained on ImageNet~\citep{Deng_Imagenet_2009} to extract visual features for each frame. HERO~\citep{Li_HERO_EMNLP20} combines 2D features from Resnet~\citep{He_ResNet_CVPR16} and 3D features from SlowFast~\citep{Feichtenhofer_SlowFast_ICCV19}.
ClipBERT~\citep{Lei_ClipBert_CVPR21} and VATT~\citep{Akbari_VATT_arxiv21} design end-to-end  frameworks without fixed visual backbone.

\paragraph{\textbf{Proxy Tasks}}
The selection or designing of the proxy tasks directly determines the model training objectives and further affects the performance on downstream tasks. 
Most pre-training works inherit the masking based tasks and matching based tasks from BERT, learning the correlation within the same modality and across different modalities.
HERO~\citep{Li_HERO_EMNLP20} and VICTOR~\citep{Lei_VICTOR_2021} design ordering tasks to explore the sequential structure of videos, which have been demonstrated beneficial to downstream tasks that rely on temporal reasoning such as videoQA.
Univl~\citep{Luo_univl_arxiv20} and VLM~\citep{Xu_VLM_ACL21} both demonstrate that masking out the whole modalities and reconstruct it based on other modalities benefits the retrieval task.

\paragraph{\textbf{Training Strategy}}
A few works develop the stage-by-stage pre-training methods instead of training the whole model in one step. For example, Univl~\citep{Luo_univl_arxiv20}, the representative of Multi-Stream transformer, trains the transformer encoder for each modality first and then the whole model with decreasing learning rate.
CBT~\citep{Sun_CBT_ECCV20} uses a curriculum learning strategy by first pre-training the visual feature extractor S3D and then jointly fine-tuning the last blocks of S3D with the visual transformer using the CBT visual loss.
Compared to training in one step, training stage-by-stage makes the pre-training progress more smoothing. 

\paragraph{\textbf{Downstream Tasks}}
To evaluate the pre-trained models, the standard approach is to transfer the pre-trained weight to other down-stream tasks. 
We compare the above methods on matching task of text-based video retrieval and generative task of video captioning. The results are shown in Tab.~\ref{tab:video-text-retrieval} and \ref{tab:video-caption} respectively. 
We divide  models according to their structure. VLM~\citep{Xu_VLM_ACL21} generally performs the best across Single-Stream models for both retrieval and captioning tasks. 
Among Multi-Stream models,  Univl~\citep{Luo_univl_arxiv20} outperforms other models generally.
VICTOR~\citep{Lei_VICTOR_2021} is not included since it is pre-trained and evaluated only on Chinese dataset.
\begin{table}[t]
    \centering
    \begin{tabular}{ccccc}
    \toprule
        Methods &  R@1 & R@5 & R@10 &  Median R  \\
        \midrule
        \textbf{Single-Stream}\\
        HERO & 16.8 & 43.4 & 57.7 & - \\
        ClipBERT & 22.0 & 46.8 & 59.9 & 6\\
        DeCEMBERT &17.5 &44.3& 58.6& 9 \\
        VLM & 28.1 &55.5 & 67.4 & 4 \\
        \hline
        \textbf{Multi-Stream}\\
        ActBERT & 8.6 & 23.4 & 33.1 & 36 \\
        Univl & 20.6 & 49.1 & 62.9 & 6 \\
    \bottomrule
    \end{tabular}
    \caption{Performance of text-based video retrieval on MSR-VTT. }
    \label{tab:video-text-retrieval}
\end{table}

\begin{table}[t]
    \centering
    \begin{tabular}{cccccc}
    \toprule
        Methods &  B-3 & B-4 & M &  R & C  \\
         \midrule
         \textbf{Single-Stream} & & & & &\\
        ViedoBERT & 6.80 & 4.04 & 11.01 & 27.50 & 0.49 \\
       
        DeCEMBERT & - & 11.92 & 20.01 & 40.22 & 0.58  \\
        VLM & 17.78 & 12.27 & 18.22 & 41.51 & 1.39\\
        \hline
        \textbf{Multi-Stream} & & & & &\\
         CBT & - & 5.12 & 12.97 & 30.44 & 0.44\\
        ActBERT & 8.66 & 5.41 & 13.30 & 30.56 & 0.65 \\
        Univl & 16.46 & 11.17 & 17.57 & 40.09 & 1.27 \\
        
    \bottomrule
    \end{tabular}
    \caption{Performance of Video Captioning on YouCookII. B, M, R, C are abbreviations of BLUE, METEOR, ROUGE, Cider. }
    \label{tab:video-caption}
\end{table}

\section{Discussion}
Pre-training has shown obvious improvements  on various Video-Language tasks compared to traditional methods. Nevertheless, the potential of transformer structure on Video-Language has not been fully explored. There still exists several challenges to be tackled. In this section, we discuss these challenges and possible future directions.
\subsection{Pre-training Dataset}
Since  transformers lack some inductive biases as CNNs, it requires large scale of datasets for pre-training.  Consequently, the quality, quantity and diversity of dataset has significant influence on the general performance of transformers. 
For the problem of quantity, the most commonly used dataset for pre-training so far is Howto100M~\citep{Miech_Howto100m_ICCV19}, which contains over 100M video-sentence pairs. Experiments on~\citep{Miech_Howto100m_ICCV19} prove that increasing the amount of training data improves the performance of variable evaluated tasks.
For the problem of quality, since large scale of manual video annotations are expensive, the corresponding captions of videos are usually generated from ASR automatically~\citep{Miech_Howto100m_ICCV19,Sun_videoBERT_ICCV19}, which inevitably introduces mistakes and misalignments to captions for corresponding video content. 
DeCEMBERT~\citep{Tang_DeCEMBERT_NAACL21} has mitigated these problems by adding extra inputs~(dense video captions~\citep{Johnson_DensCap_CVPR16}) and adjusting the training objective. 
For the problem of diversity, pre-training dataset used in VideoBERT~\citep{Sun_videoBERT_ICCV19} focuses on cooking domain. Videos of Alivol-10M~\citep{Lei_VICTOR_2021} come from E-commerce website. Videos of Howto100M~\citep{Miech_Howto100m_ICCV19} are crawled from YouTube. 
These pre-training datasets are mainly from a single domain and inevitably have domain gaps with various downstream datasets, which has been demonstrated to be harmful to the performance of pre-trained models. 
On this topic, Zhou et al.~\citep{Zhou_CUPID_arxiv21} proved that pre-training on a considerably small subset of domain-focused data can effectively close the source-target domain gap and achieve significant performance gain. Similar conclusion can be found in HERO~\citep{Li_HERO_EMNLP20} that domain gap of finetuning and pre-training can not be eliminated by data volume. 
In conclusion, although a lot of explorations have been done, there is still a long way to go in order to improve the quantity, quality and diversity of pre-training datasets.

\subsection{Video-Language Transformer Designs}


Existing works mostly follow the paradigm from NLP domain and make adjustments to adapt to Video-Language processing, which includes using multi-stream structure to meet the needs of multimodal input, designing reordering proxy tasks to exploit the sequential structure of videos, and adding audio modality as supplementary information.
Although the results of these applications are quite encouraging, current methods require further intuition to better match the Video-Language tasks. 
Firstly, how to deal with the visual backbone properly remains unsolved. Existing works either apply a independently trained visual backbone to extract video features~\citep{Xie_S3D_ECCV18} or train a model that includes 2D CNN backbone in the end-to-end manner~\citep{Lei_ClipBert_CVPR21}. 
The first type of approach not only leads to domain gap between feature extraction and pre-training, but also hinders model improvement due to the loss of fine-grained visual information. The other type of approach tends to lose the temporal information in the video.
Secondly, standard evaluation of Video-Language pre-training is an urgent need for sustainable development in this field. So far, different models are evaluated on different downstream tasks/datasets with different detailed settings, which is unfair to compare their performance. A unified benchmark is needed to evaluate different pre-training methods, such as GLUE~\citep{Wang_Glue_EMNLP18} in NLP. VALUE~\citep{Li_VALUE_arxiv21} has proposed an evaluation benchmark that covers 11 datasets over 3 popular tasks including retrieval,caption and videoQA, but it has not yet been popularized. 

Another promising direction is to improve the generalization ability of pre-trained models. As a collection of multiple modalities, a video contains more than semantic information. 
For example, ActBERT~\citep{Zhu_ActBERT_CVPR20} uses fine-grained object regions of videos, VATT~\citep{Akbari_VATT_arxiv21} explores the inner relationship of frame sequence, audio and sentence. 
We believe that there exist more clues that can mined from videos, such as scene information, character information. How to make use of these information and transfer to more downstream tasks are promising future directions. 
On the basis of multiple input, video analysis should not be limited to analysis of general semantics.
Tasks related to image, audio, and text modalities are  expected to be covered by a comprehensive model.
What's more, we notice that existing works mostly focus on the domains of activity, films, and TV shows. Other domains such as medical field, surveillance recordings have a lot of potential applications as well.

\subsection{Transformer Efficiency}

A well-known concern of transformer is the efficiency problems of quadratic time and memory complexity,  which hinders transformer's scalability in practice. 
As mentioned in~\citep{Tay_EfficientTransformer_arxiv20}, model efficiency refers to both  memory footprint and computation cost. For the memory efficient transformers, Lee et al.~( \cite{Lee_ParameterShare_ICLR21} use weight sharing across layers and modalities to reduce overall model size. Similar idea is originated from Universal Transformers~\citep{Dehghani_UniversalTransformer_ICLR19} and Albert~\citep{Lan_ALbert_ICLR20}.
For the computation efficient transformers, Michel et al.~\citep{Michel_CutHead_NeurIPS19} remove some heads at test time without impacting performance. \cite{Prasanna_CutBERT_EMNLP20}  also reduce the computation cost by pruning and decomposing original transformer structure.

In summary, studies of efficient transformers are mainly from NLP domain, which focus more on handling longer sequence input. Video naturally conveys more information than pure text. Video-Language processing requires deeper model structure, larger parameters and thus has higher requirements for hardware and computation. 
\section{Conclusion}
Pre-training has become a popular approach in NLP  and has been further applied in vision tasks. Comparing to other vision-language pre-training works, less pre-training works reported in the Video-Language area. We therefore conduct a comprehensive overview of pre-training methods for Video-Language processing in this paper.
This survey first reviews the background knowledge related to transformer, then summarizes the  pre-training and finetuning process of Video-Language learning by introducing the common proxy tasks and downstream tasks respectively. Furthermore, we describe commonly used video datasets according to their scale, annotation type etc. 
We also summarize the state of the art transformer models for Video-Language learning, highlight their key strength and compare their down-stream performance.
Finally, we conclude the paper with discussions of the current challenges and possible future research directions.








\printcredits

\bibliographystyle{cas-model2-names}

\bibliography{main}



\end{document}